\documentclass[journal,12pt,onecolumn,draftclsnofoot]{IEEEtran}
\usepackage{amsmath,epsfig}
\usepackage{amssymb}
\usepackage{graphicx}
\usepackage{epstopdf}
\usepackage{url}
\usepackage{float}
\usepackage[noadjust]{cite}

\begin{document}

\title{ Recurrent Neural Network Training with Preconditioned Stochastic Gradient Descent}
\author{ Xi-Lin Li, lixilinx@gmail.com  }

\maketitle

\begin{abstract}
This paper studies the performance of a recently proposed preconditioned stochastic gradient descent (PSGD) algorithm on recurrent neural network (RNN) training. PSGD adaptively estimates a preconditioner to accelerate gradient descent, and is designed to be simple, general and easy to use, as stochastic gradient descent (SGD). RNNs, especially the ones requiring extremely long term memories, are difficult to train. We have tested PSGD on a set of synthetic pathological RNN learning problems and the real world MNIST handwritten digit recognition task. Experimental results suggest that PSGD is able to achieve highly competitive performance without using any trick like preprocessing, pretraining or parameter tweaking. 
\end{abstract}


\section{Introduction}

In its simplest form, a standard one dimensional recurrent neural network (RNN) can be defined as
\begin{equation}
	\pmb x(t) = \phi\left( \pmb W_1 \left[ \begin{array}{c}
	\pmb u(t) \\
	\pmb x(t-1) \\
	1
	\end{array} \right] \right), \quad \pmb y(t) = \pmb W_2 \left[ \begin{array}{c}
	\pmb x(t) \\
	1
	\end{array} \right],
\end{equation}
where $t$ is a discrete time index, $\phi$ an element-wise sigmoid function, $\pmb u$ the input sequence, $\pmb x$ the hidden state sequence, $\pmb y$ the output sequence, and $\pmb W_1$ and $\pmb W_2$ are two weight matrices with proper dimensions. RNN is a powerful tool for sequence modeling, and its gradient can be conveniently calculated, e.g., via backpropagation through time (BPTT) \cite{bptt}. Unfortunately, learning RNN  turns out to be extremely difficult when it is used to solve problems requiring long term memories \cite{difficulty_rnn, Martens2012_hessian_free, lstm}. Exploding and vanishing gradients, especially the latter one, are suspected to be the causes. Hence long and short term memory (LSTM) and its variants \cite{lstm, lstm_variant} are invented to overcome the vanishing gradient issue mainly by the use of forgetting gates. However, as a specially modified model, LSTM may still fail to solve certain problems that are suitable for its architecture, e.g., finding XOR relationship between two binary symbols with a long lag. Furthermore, as a more complicated model, it does not necessarily always outperform the standard RNN model on certain natural problems as reported in \cite{difficulty_rnn, Martens2012_hessian_free}. Another way to address vanishing gradient is to directly penalize RNN connections encouraging vanishing gradients \cite{difficulty_rnn}. However, as an ad hoc method, its impact on the convergence and performance of RNN training is unclear. One important discovery made in \cite{Martens2012_hessian_free} is that RNN requiring long term memories can be trained using Hessian free optimization, a conjugate gradient (CG) method tailored for neural network training with the use of a backpropagation like procedure for curvature matrix-vector product evaluation \cite{Schraudolph02}. However, due to its use of line search, Hessian free optimization requires a large mini-batch size for gradient and cost function evaluations, making it computationally demanding for problems with large training sample sizes. 

Recently, a preconditioned stochastic gradient descent (PSGD) algorithm is proposed in \cite{Li15}. It is a simple and general procedure to upgrade a stochastic gradient descent (SGD) algorithm to a second-order algorithm by exploiting the curvature information extracted exclusively from noisy stochastic gradients. It is virtually tuning free, and applicable equally well to both convex and non-convex problems, a striking difference from many optimization algorithms, including the Hessian free one, which assume  positive definite Hessian matrices at least for their derivations. Naturally, we are curious about its performance on RNN training, especially on those challenging pathological synthetic problems since they are effectively impossible for SGD \cite{lstm, Martens2012_hessian_free}. Our results suggest that although the issue of exploding and vanishing gradients arises naturally in RNN, efficient learning is still possible when the gradients are properly preconditioned. Experimental results on the MNIST handwritten digit recognition task suggest that preconditioning helps to improve convergence as well even when no long term memory is required.    

\section{PSGD and RNN}

\subsection{PSGD}

We briefly summarize the PSGD theory in \cite{Li15}. Let us consider the minimization of cost function,
\begin{equation}
	f(\pmb \theta) = E\left[ \ell(\pmb \theta, \pmb z) \right],
\end{equation}
where $\pmb\theta$ is a parameter vector to be optimized, $\pmb z$ is a random
vector, $\ell$ is a loss function, and $E$ takes expectation over $\pmb z$. At the $k$th iteration of PSGD, we evaluate two stochastic gradients over the same randomly drawn samples: the original gradient $\pmb g_k$ at point $\pmb\theta=\pmb \theta_k$, and a perturbed gradient $\tilde{\pmb g}_k$ at point $\pmb\theta=\pmb \theta_k + \delta \pmb \theta_k$, where $\delta \pmb \theta_k$ is a tiny random vector. By introducing gradient perturbation as $\delta \pmb g_k = \tilde{\pmb g}_k - \pmb g_k$, a positive definite preconditioner, $\pmb P_k$, can be pursued by minimizing criterion 
\begin{equation}\label{criterion}
	E\left[ \delta \pmb g_k^T \pmb P_k \delta \pmb g_k + \delta \pmb \theta_k^T \pmb P_k^{-1}\delta \pmb \theta_k\right],
\end{equation}
where $E$ takes expectation over random vector $\delta \pmb \theta_k$. Under mild conditions, such a $\pmb P_k$ exists and is unique \cite{Li15}. 
As a result, the PSGD learning rule is written as,
\begin{equation}
	\pmb\theta_{k+1} = \pmb\theta_{k} - \mu \pmb P_k \pmb g_k, 
\end{equation}   
where $0<\mu<1$ is a normalized step size. The preconditioner can be conveniently estimated using stochastic relative (natural) gradient descent with mini-batch size $1$. 

The rationality of PSGD is that minimizing criterion (\ref{criterion}) leads to a preconditioner scaling the gradient such that the amplitude of $\pmb P_k\delta \pmb g_k$ approximately matches that of $\delta \pmb \theta_k$ as $\pmb P_k E[\delta \pmb g_k \delta \pmb g_k^T] \pmb P_k = E[\delta \pmb \theta_k\delta \pmb \theta_k^T]$. When gradient noise vanishes, we have $\pmb P_k \delta \pmb g_k \delta \pmb g_k^T \pmb P_k =  \delta \pmb \theta_k\delta \pmb \theta_k^T$, a relationship comparable to  $\pmb H_k^{-1} \delta \pmb g_k \delta \pmb g_k^T \pmb H_k^{-1} =  \delta \pmb \theta_k\delta \pmb \theta_k^T$, where $\pmb H_k$ is the Hessian at $\pmb\theta=\pmb \theta_k$. Hence PSGD can be regarded as a stochastic version of the deterministic Newton method when $\pmb P_k$ converges to $\pmb H_k^{-1}$ and $\mu=1$. But unlike the Newton method, PSGD applies equally well to non-convex optimizations since $\pmb P_k$ can be chosen to be positive definite even when $\pmb H_k$ is indefinite. 

In the context of RNN training, $\pmb P_k$ damps exploding gradients and amplifies vanishing gradients by trying to match the scales of vectors $\pmb P_k\delta \pmb g_k$ and $\delta \pmb \theta_k$. In this way, a single preconditioner solves both the exploding and vanishing gradient issues in learning RNN, while conventionally, several different strategies are developed and combined to fix these two issues, e.g., gradient clipping, using penalty term to discourage vanishing gradient, forgetting gate, etc..     

\subsection{Application to RNN Training}

\subsubsection{Dense preconditioner}

It is straightforward to apply PSGD to RNN training by stacking all the elements in $\pmb W_1$ and $\pmb W_2$ to form a single coefficient vector $\pmb \theta$. The resultant preconditioner has no sparsity. Hence, such a brutal force solution is practical only for small scale problems with up to thousands of parameters to learn.   

\subsubsection{Preconditioner with sparse structures}

For large scale problems, it is necessary to enforce certain sparse structures on the preconditioner so that it can be stored and manipulated on computers. Supposing the dimensions of $\pmb u$, $\pmb x$ and $\pmb y$ are $N_u$, $N_x$ and $N_y$ respectively, one example is to enforce $\pmb P$ to have form
\begin{equation}
	\pmb P = \left( \pmb P_2 \otimes \pmb P_1 \right) \oplus \left( \pmb P_4 \otimes \pmb P_3 \right),
\end{equation}
where the dimensions of positive definite matrices $\pmb P_1$, $\pmb P_2$, $\pmb P_3$, and $\pmb P_4$ are $N_x$, $N_u+N_x+1$, $N_y$ and $N_x+1$ respectively, and $\otimes$ and $\oplus$ denote Kronecker product and direct sum respectively. Algorithms for learning these $\pmb P_i$, $1\le i\le 4$, are detailed in \cite{Li15} as well. We mainly study the performance of PSGD with sparse preconditioner due to its better scalability with respect to problem sizes. 

\section{Experimental Results}	

We consider a real world handwritten digit recognition problem \cite{mnist}, and a set of pathological synthetic problems originally proposed in \cite{lstm} and restudied in \cite{Martens2012_hessian_free, difficulty_rnn}. Details of these pathological synthetic problems can be found in \cite{lstm} and the supplement of \cite{Martens2012_hessian_free}. For continuous problems (outputs are continuous), mean squared error (MSE) loss is used, and for discrete problems (outputs are discrete), cross entropy loss is used. The same parameter settings as in \cite{Li15} are used for PSGD, and no problem-specific hand tweaking is made. Specifically, the preconditioner is initialized to identity matrix, and then updated using stochastic relative gradient descent with mini-batch size $1$, step size $0.01$ and sampling $\delta\pmb \theta$ from Gaussian distribution $\mathcal{N}(0, {\rm eps})$ element-wisely, where ${\rm eps}=2^{-52}$ is the accuracy in double precision. The recurrent matrix of RNN is initialized to a random orthogonal matrix such that neither exploding nor vanishing gradient issue is severe at the beginning, loosely comparable to setting large initial biases in the forgetting gates of LSTM \cite{lstm_variant}. Other non-recurrent weights are element-wisely initialized to small random numbers drawn from normal distribution. Mini-batch size $100$ and step size $0.01$ are used for RNN training. Program code written in Matlab and supplemental materials revealing more detailed experimental results can be found at \url{https://sites.google.com/site/lixilinx/home/psgd}. 

\subsection{Experiment 1: PSGD vs. SGD} 

We consider the addition problem from \cite{lstm} where a RNN is trained to predict the sum of a pair of marked, but randomly located, continuous random numbers in a sequence. For SGD, clipped stochastic gradient with clipping threshold $1$ is used to address the exploding gradient issue. SGD seldom succeeds on this problem when the sequence length is no less than $100$. To make the problem easier, sequences with length uniformly distributed in range $[50, 100]$ are used for training, hoping that SGD can learn the desired patterns from shorter sequences and then generalize them to longer ones. Fig.~1 shows three learning curves for three algorithms using the same initial guess and step size: SGD, PSGD with a sparse preconditioner, and PSGD with a dense preconditioner. Clearly, PSGD with a dense preconditioner converges the fastest. The sparse preconditioner helps a lot as well, despite its simplicity. SGD converges the slowest.              

\begin{figure}[H]
	\centering
	\includegraphics[width=0.5\textwidth]{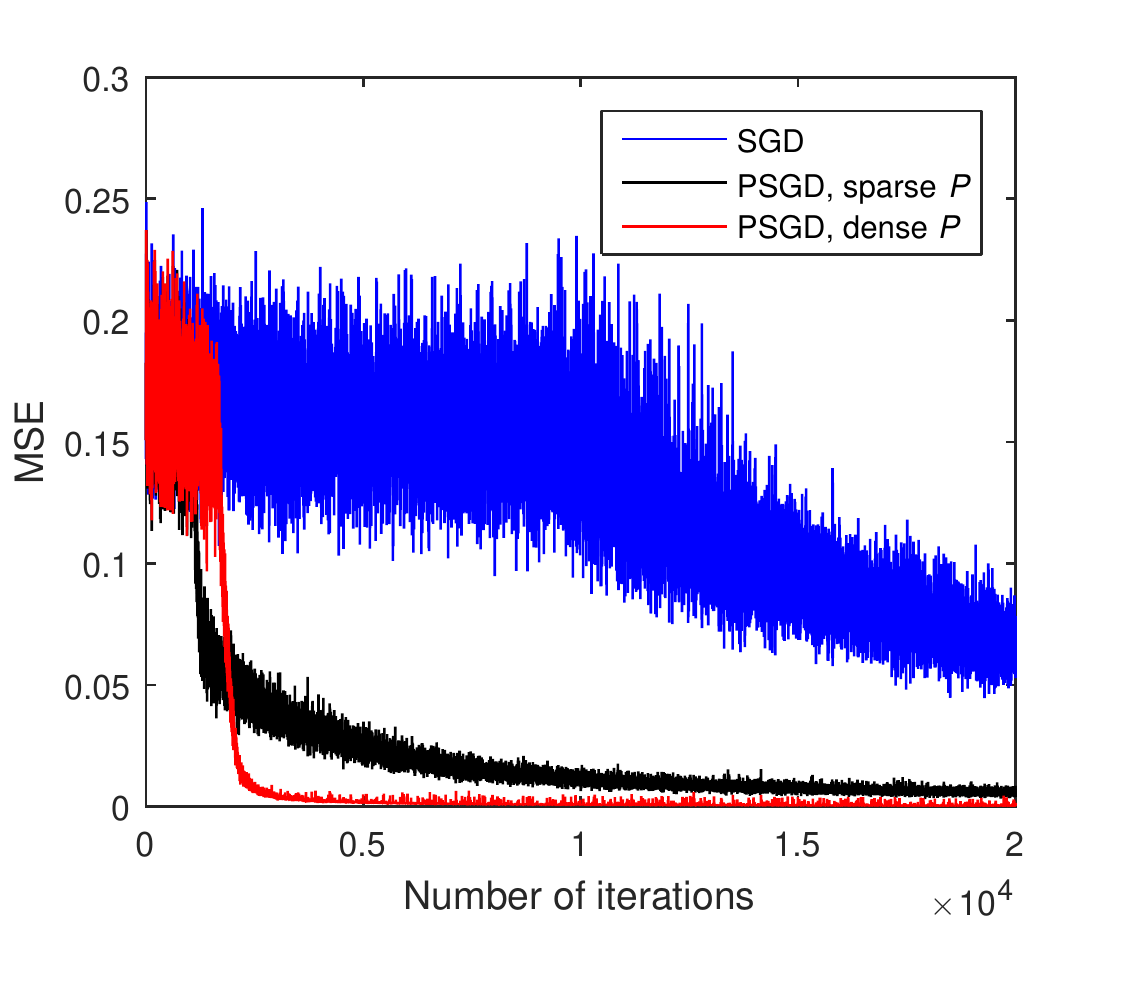}\\
	\caption{ Convergence curves of SGD and PSGD for the addition problem. The hidden layer has $50$ neurons.  }
\end{figure}

\subsection{Experiment 2: Performance on Pathological Synthetic Problems}

We consider the four groups of pathological synthetic problems in \cite{lstm}. The first group includes the addition, multiplication, and XOR problems; the second group includes the $2$-bit and $3$-bit temporal order problems; the third group only has the random permutation problem; and the fourth group are the $5$-bit and $20$-bit noiseless memorization problems. Totally we have eight problems. In the addition and multiplication problems, RNN needs to memorize continuous random numbers with certain precision for many steps. In the $2$-bit and $3$-bit temporal order problems, RNN needs to memorize two and three widely separated binary bits and their order, respectively. The XOR problem challenges both RNN and LSTM training since this problem cannot be decomposed into smaller ones. In the random permutation problem, RNN is taught to predict random unpredictable symbols, except the one at the end of sequence, leading to extremely noisy gradients. On the contrary, all symbols in the $5$-bit and $20$-bit memorization problems, except those information carrying bits, can be trivially predicted, but are not task related, thus diluting the importance of task related gradient components. 

We follow the experimental configurations in \cite{Martens2012_hessian_free, lstm} so that the results can be compared. The results reported in \cite{difficulty_rnn} could be biased because according to the descriptions in \cite{difficulty_rnn}, for most problems, RNN is trained on sequences with length uniformly distributed in range $[50, 200]$. This considerably facilitates the training since RNN has chances to learn the desired patterns from short sequences and then to generalize them to long ones, as shown in Experiment 1. We follow the configurations in \cite{Martens2012_hessian_free, lstm} to ensure that there is no short time lag training exemplar to facilitate learning. 

Among these eight problems, the $5$-bit memorization problem is special in the way that it only has $32$ distinct input sequences. Hence we set its mini-batch size to $32$. Then the gradient is exact, no longer stochastic. PSGD applies to deterministic optimization as well, but extra cares need to be taken to prevent the arising of an ill-conditioned Hessian since PSGD is essentially a second-order optimization algorithm. Note that the cross entropy loss is invariant to the sum of elements in $\pmb y$. Thus $\pmb W_2$ only needs to have $(N_y-1)(N_x+1)$ degrees of freedom. Its extra $N_x+1$ degrees of freedom cause singular Hessian all over the parameter space. We remove those extra $N_x+1$ degrees of freedom in $\pmb W_2$ by constraining all its columns having zero sum. We would like to point out that gradient noise in stochastic gradient naturally regularizes the preconditioner estimation as shown in \cite{Li15}. Hence we have no need to remove those extra $N_x+1$ degrees of freedom in $\pmb W_2$ for the other five discrete problems.

Only the PSGD with sparse preconditioner is tested. For each problem, four sequence lengths, $30$, $50$, $100$ and $200$, are considered. For each problem with each sequence length, five independent runs starting from different random initial guesses are carried out. A run is said to be failed when it fails to converge within the maximum allowed number of iterations, which is set to $10^5$ for PSGD. Table I summarizes the failure rate results. Note that RNN training may take a long time. Hence, we have not finished all five runs for a few test cases due to limited resources. 

\begin{table}[H]
	\centering
	\caption{ PSGD's failure rate shown as (number of failed runs)/(number of total runs) on eight pathological problems  }\label{table1}
	\begin{tabular}{|c|c|c|c|c|}
		\hline
		&  30 & 50 & 100 & 200  \\ \hline
		Addition & 0/5 & 0/5 & 0/5 & 2/5 \\ \hline
		Multiplication & 0/5 & 0/5 & 0/5 & 0/5 \\ \hline
		XOR & 0/5 & 0/5 & 3/5 & 1/1 \\ \hline
		$2$-bit temporal order & 0/5 & 0/5 & 0/5 & 3/4 \\ \hline
		$3$-bit temporal order & 0/5 & 0/5 & 0/5 & 2/3 \\ \hline
		Random permutation & 0/5 & 0/5 & 0/5 & 0/5  \\ \hline
		$5$-bit memorization & 0/5 & 0/5 & 0/5 & 0/5  \\ \hline
		$20$-bit memorization & 0/5 & 0/5 & 0/5 & 0/5  \\ \hline
	\end{tabular}
\end{table}

We compare our results with the ones reported in \cite{Martens2012_hessian_free}. Since only a few runs are carried out, neither the result here nor the one in \cite{Martens2012_hessian_free} has statistical significance. Thus we would like to compare the maximum sequence length that an algorithm can handle without failure. This criterion favors the results reported in \cite{Martens2012_hessian_free} as for each problem with each sequence length, only four runs are done there, while PSGD has five runs. These results are summarized in Table II. From Table II, we observe that PSGD outperforms Hessian-free optimization with Tikhonov damping on the multiplication, XOR, $2$-bit temporal order, $5$-bit memorization, and $20$-bit memorization problems. PSGD outperforms Hessian-free optimization with structural damping on the multiplication, $2$-bit temporal order, random permutation, and $20$-bit memorization problems. Overall speaking, PSGD outperforms Hessian-free optimization with either Tikhonov damping or structural damping, and its performances are no worse than the best ones achieved by both versions of Hessian-free optimization. 

\begin{table}[H]
	\centering
	\caption{ Maximum sequence length without observing failure   }\label{table1}
	\begin{tabular}{|c|c|c|c|}
		\hline
		&  HF, Tikhonov & HF, structural & PSGD   \\ \hline
		Addition & 100 & 100 & 100 \\ \hline
		Multiplication & 100 & 100 & 200 \\ \hline
		XOR & 30 & 50 & 50 \\ \hline
		$2$-bit temporal order & 50 & 50 & 100 \\ \hline
		$3$-bit temporal order & 100 & 100 & 100 \\ \hline
		Random permutation & 200 & 100 & 200 \\ \hline
		$5$-bit memorization & $<30$ & 200 & 200 \\ \hline
		$20$-bit memorization & $30$ & 100 & 200 \\ \hline
	\end{tabular}
\end{table}

\subsection{Experiment 3: MNIST Handwritten Digit Recognition}

Not every practical RNN learning problem is as pathological as the above studied synthetic problems. Still, PSGD could take nontrivial advantages over SGD such as faster and better convergence even when no long term memory is required. Here, the classic MNIST handwritten digit recognition task is considered \cite{mnist}. The original $28\times 28$ images are zero padded to $32\times 32$ ones. Fig.~2 shows the architecture of a small but deep two dimensional RNN used to recognize the zero padded images. No long term memory is required as either dimension only requires eight steps of back propagation.  

\begin{figure}[H]
	\centering
	\includegraphics[width=0.5\textwidth]{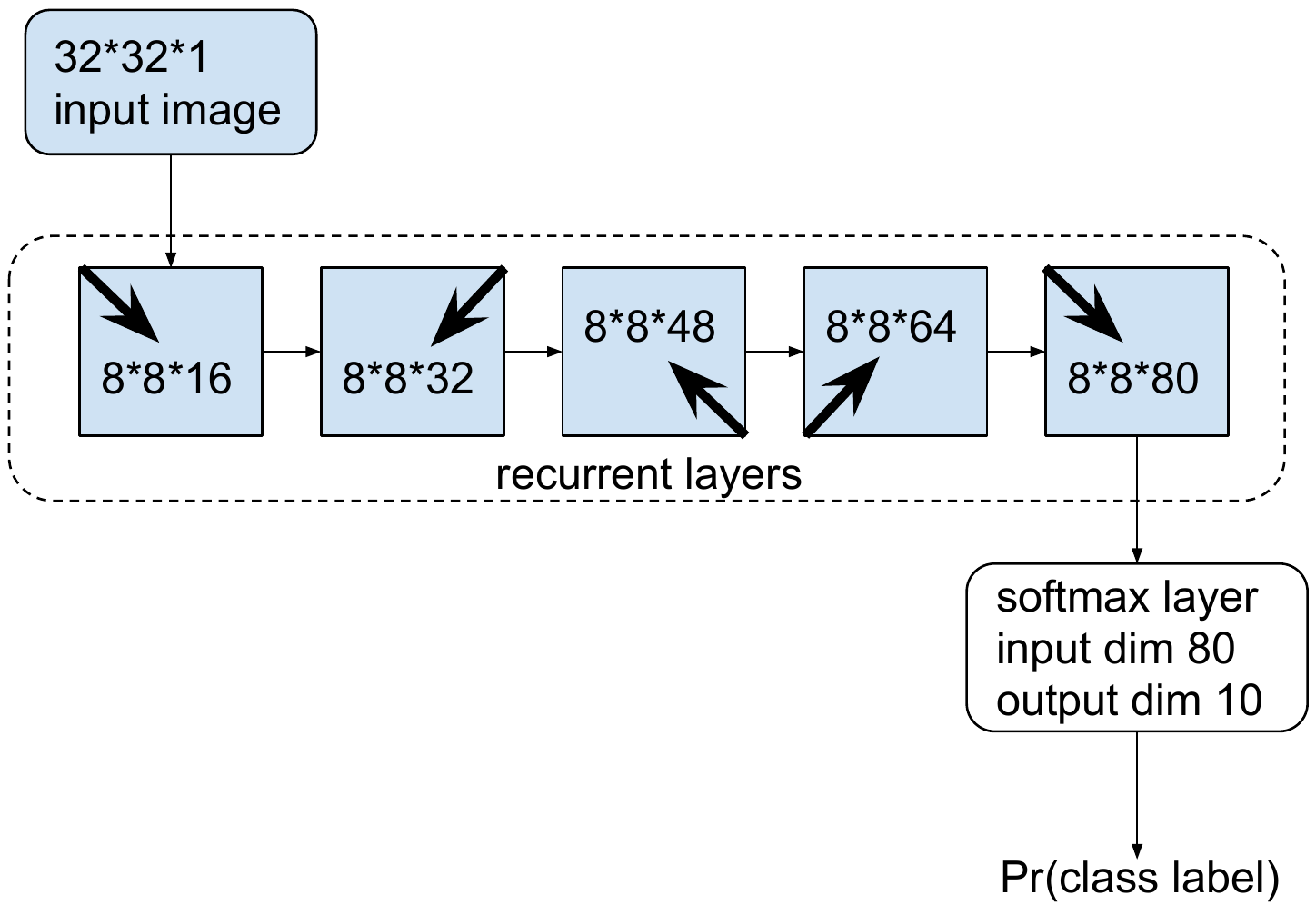}\\
	\caption{ The two dimensional RNN used in MNIST handwritten digit recognition. Neuron in the first recurrent layer has a receptive field of $4\times 4$ pixels without overlapping. Feature dimensions in the five RNN layers are $16$, $32$, $48$, $64$, and $80$, respectively. Boldface arrow points the propagation direction of state variables in each recurrent layer. The last layer is a softmax function with the last state of the last recurrent layer as its input.    } 
\end{figure}

Both SGD and PSGD start from the same random initial guess, and use the same step size and mini-batch size. PSGD uses layer-wise Kronecker product preconditioner. No preprocessing, pretraining or artificially distorted version of the original training samples is used. Fig.~3 plots the test error rate convergence curves. Here, the test error rate is the ratio of the number of misclassified testing samples to the total number of testing samples. From Fig.~3, one observes that PSGD always converges faster and better than SGD. It is interesting to compare the test error rates here with that listed on \cite{mnist} achieved by convolutional neural networks without using distorted version of the original training samples. Here, SGD and PSGD converge to test error rates $0.9\%$ and $0.6\%$, respectively. They are comparable to the ones listed on \cite{mnist} achieved using convolutional neural networks without and with pretraining, respectively.         

\begin{figure}[H]
	\centering
	\includegraphics[width=0.5\textwidth]{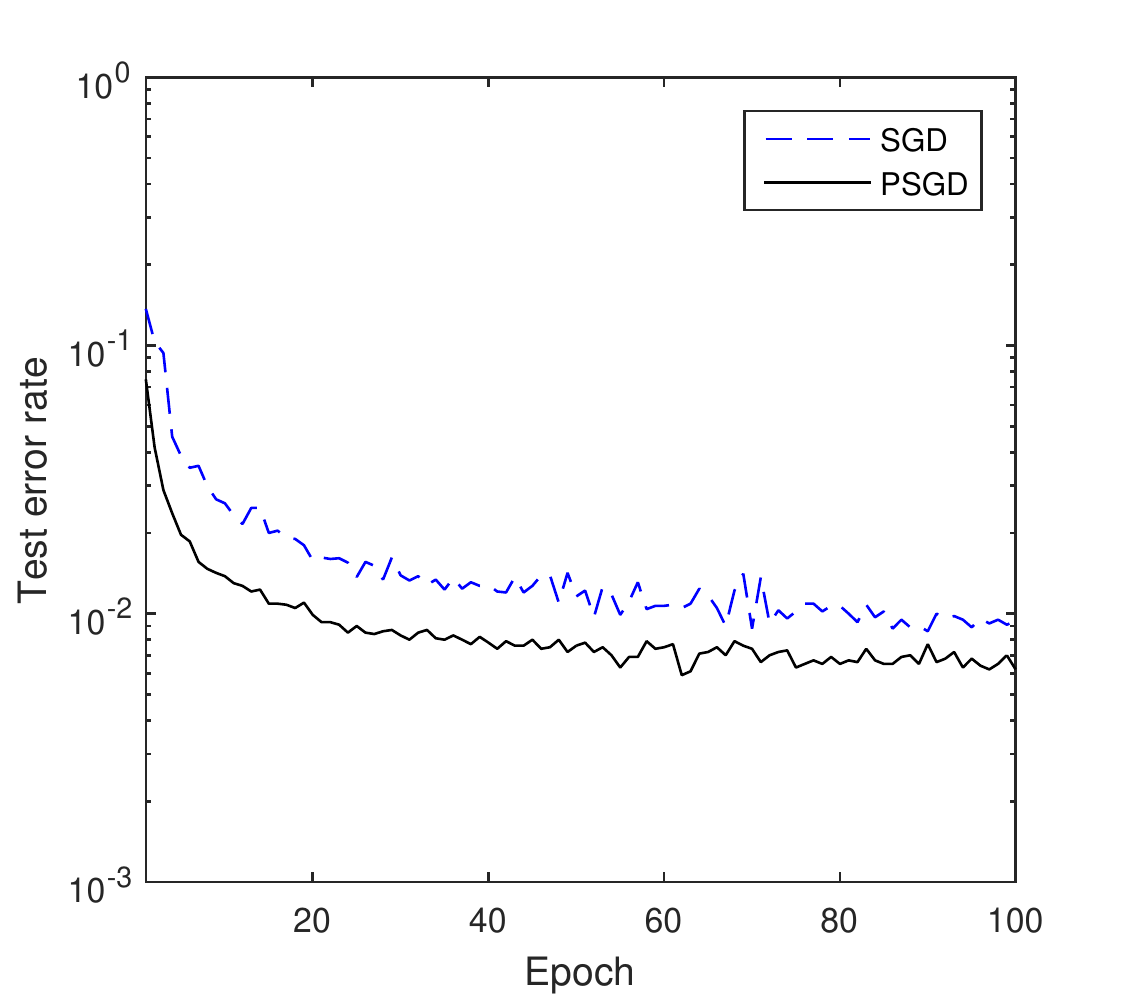}\\
	\caption{ Test error rate convergence curves for the MNIST example.  } 
\end{figure}

\section{Conclusions and Discussions}

Preconditioned stochastic gradient descent (PSGD) is a general and simple learning algorithm, and requires little tuning effort. We have tested PSGD on eight pathological synthetic recurrent neural network (RNN) training problems. Although these problems may fail stochastic gradient descent (SGD) miserably, PSGD works quite well on them, and even could outperform Hessian-free optimization, a significantly more complicated algorithm than both SGD and PSGD. While SGD is workable for many practical problems without requiring long term memory, PSGD still provides nontrivial advantages over it such as faster and better convergence as demonstrated in the MNIST handwritten digit recognition example.    

Unlike many traditional second-order optimization algorithms which assume positive definite Hessian, PSGD is designed for both convex and non-convex optimizations. This might explains its superior performance even its implementation is just slightly more complicated than SGD. PSGD works well with small mini-batch sizes to reduce computational complexity due to its inherent ability to damp gradient noise naturally, while many off-the-shelf algorithms require a large mini-batch size for accurate gradient and cost function evaluations to facilitate line search. Furthermore, PSGD is easier to use since its step size is normalized, saving the trouble of step size selection by either hand tweaking or using step size searching algorithms. Its preconditioner can have flexible forms, providing trade off room between performance and complexity. These properties make PSGD an attractive alternative to SGD and many other stochastic optimization algorithms.


\begin{thebibliography}{1}

\bibitem{bptt}
P.~J.~Werbos, ``Backpropagation through time: what it does and how to do it,'' \emph{Proc. IEEE}, vol.~78, no.~10, pp.~1550--1560, Oct. 1990. 

\bibitem{difficulty_rnn}	
R.~Pascanu, T.~Mikolov, and Y.~Bengio, ``On the difficulty of training recurrent neural networks,'' arXiv:1211.5063, 2012. 

\bibitem{Martens2012_hessian_free}
J.~Martens and I.~Sutskever, ``Learning recurrent
neural networks with Hessian-free optimization,'' In \emph{Proc. of the 28th ICML}, 2011.

\bibitem{lstm}
S.~Hochreiter and J.~Schmidhuber, ``Long short-term memory,'' \emph{Neural Computation}, vol.~9, no.8, pp.~1735--1780, 1997.

\bibitem{lstm_variant}
K.~Greff, R.~K.~Srivastava, J.~Koutník, B.~R.~Steunebrink, and J.~Schmidhuber, ``LSTM: a search space odyssey,'' arXiv:1503.04069, 2015. 

\bibitem{Schraudolph02}
N.~Schraudolph, ``Fast curvature matrix-vector products for
second-order gradient descent,'' \emph{Neural Computation}, vol.~14, no.~7, pp.~1723--1738, 2002.

\bibitem{Li15}
X.-L.~Li, ``Preconditioned stochastic gradient descent,'' arXiv:1512.04202, 2015. 

\bibitem{mnist}
Y.~LeCun, C.~Cortes, and C.~J.~C.~Burges, \emph{THE MNIST DATABASE}. Retrieved from \url{http://yann.lecun.com/exdb/mnist/}.

\end{thebibliography}
\end{document}